\documentclass[]{bytedance_seed}

\usepackage[toc,page,header]{appendix}

\usepackage{minitoc}
\usepackage{amssymb}
\usepackage{pifont} 
\def\gou{\ding{51}}
\def\cha{\ding{55}}
\usepackage{algorithm} 
\usepackage{algpseudocode} 
\usepackage{makecell}
\usepackage[table]{xcolor}

\usepackage{cleveref}
\crefname{equation}{Eq}{Eqs}
\crefname{table}{Table}{Tables}
\crefname{figure}{Figure}{Figures}
\crefname{algorithm}{Algorithm}{Algorithms}
\crefname{appendix}{Appendix}{Appendix}
\crefname{section}{Section}{Section}

\def \ie{{\textit{i.e.}}}

\title{FARMER: Flow AutoRegressive Transformer over Pixels}
\author[1,3]{Guangting Zheng}
\author[1,4]{Qinyu Zhao}
\author[1]{Tao Yang}
\author[1]{Fei Xiao} 
\author[2]{Zhijie Lin}
\author[1]{\\[1mm]Jie Wu}
\author[5]{Jiajun Deng}
\author[3]{Yanyong Zhang}
\author[1\dagger]{Rui Zhu}

\affiliation[1]{ByteDance Seed China}
\affiliation[2]{ByteDance Seed Singapore}
\affiliation[3]{University of Science and Technology of China}
\affiliation[4]{Australian National University}
\affiliation[5]{National University of Singapore}

\contribution[\dagger]{Project lead}

\abstract{
Directly modeling the explicit likelihood of the raw data distribution is key topic in the machine learning area, which achieves the scaling successes in Large Language Models by autoregressive modeling. However, continuous AR modeling over visual pixel data suffer from extremely long sequences and high-dimensional spaces. In this paper, we present FARMER, a novel end-to-end generative framework that unifies Normalizing Flows (NF) and Autoregressive (AR) models for tractable likelihood estimation and high-quality image synthesis directly from raw pixels. FARMER employs an invertible autoregressive flow to transform images into latent sequences, whose distribution is modeled implicitly by an autoregressive model. To address the redundancy and complexity in pixel-level modeling, we propose a self-supervised dimension reduction scheme that partitions NF latent channels into informative and redundant groups, enabling more effective and efficient AR modeling. Furthermore, we design a one-step distillation scheme to significantly accelerate inference speed and introduce a resampling-based classifier-free guidance algorithm to boost image generation quality. Extensive experiments demonstrate that FARMER achieves competitive performance compared to existing pixel-based generative models while providing exact likelihoods and scalable training.  
}

\date{\today}
\correspondence{\email{zhurui.kim@bytedance.com}}

\begin{document}
\maketitle


\section{Introduction}
\begin{figure}[t]
    \centering
    \begin{subfigure}[t]{0.28\linewidth}
        \centering
        \includegraphics[height=7cm]{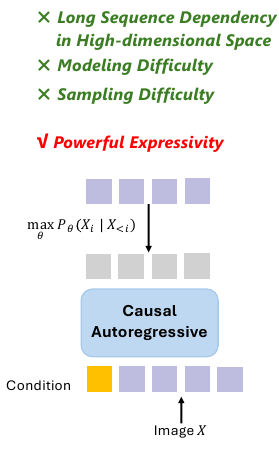}
        \caption{Pixel Autoregressive models}
        \label{fig:intro_a}
    \end{subfigure}
    \begin{subfigure}[t]{0.26\linewidth}
        \centering
        \includegraphics[height=7cm]{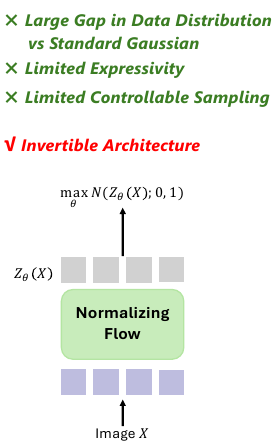}
        \caption{Normalizing Flow}
        \label{fig:intro_b}
    \end{subfigure}
    \begin{subfigure}[t]{0.44\linewidth}
        \centering
        \includegraphics[height=7cm]{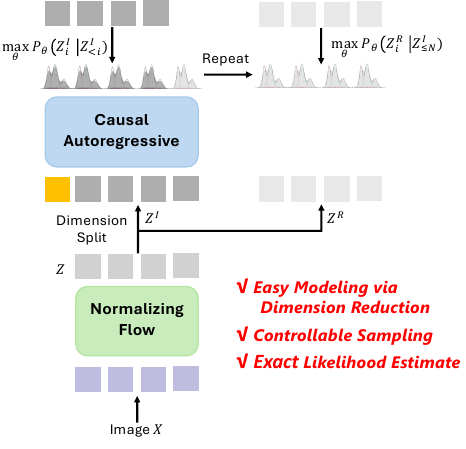}
        \caption{Flow Autoregressive Transformer (FARMER)}
        \label{fig:intro_c}
    \end{subfigure}
    \caption{
    Autoregressive (AR) models offer strong expressivity but struggle with pixel modeling and sampling due to the long sequences required for high-resolution images. Normalizing flows (NFs) employ invertible mappings to transform complex image distributions to a standard Gaussian, but the substantial gap between two distributions leads to degraded sampling quality. FARMER unifies NF and AR within a single framework, using the NF component to transform images into latent sequences, whose distribution is implicitly modeled by the AR component for easier modeling and controllable sampling. Furthermore, FARMER adopts a self-supervised dimension reduction method to partition NF latent channels into distinct groups, making AR modeling feasible and scalable.}
    \label{fig:intro}
\end{figure}

Explicitly modeling a normalized likelihood $\mathbf{P}(x)$ over the high-dimensional data distribution is challenging. 
Popular generative paradigms such as Variational Autoencoders (VAEs), Generative Adversarial Networks (GANs), and diffusion/score-based models do not provide  tractable likelihoods—VAEs optimize a lower bound, GANs learn implicit generators without likelihoods, and diffusion/score-based models offer likelihoods only via variational bounds or costly numerical estimation by probability-flow ODE. 
In contrast, Autoregressive (AR) models directly factorize sequence likelihoods via the chain rule and lead to the scaling successes of Large Language Models~\cite{gpt4,gemini,qwen,llama,gemma}. However, modeling the likelihood over continuous, high-dimensional image pixels remains notably challenging compared to  the discrete text.
Continuous AR over visual pixels has been explored for years—from convolutional PixelRNN/PixelCNN~\cite{van2016conditional,van2016pixel} to  Image Transformer~\cite{parmar2018image} and iGPT~\cite{chen2020generative}. 
Despite these efforts, continuous AR suffers from extremely long sequences, making training and sampling costly and brittle to long-range dependencies. 
This gap motivates revisiting how we parameterize continuous densities over high-dimensional pixel spaces and how we couple them with scalable sequence models.

At the same time, Normalizing Flow (NF)~\cite{kolesnikov2024jet,zhai2024normalizing,gu2025starflow} has seen a resurgence for image generation. By providing exact likelihoods via invertible and differentiable mappings, 
NF offers an attractive route for revitalizing continuous AR modeling and a principled latent representation. 
For instance, JetFormer~\cite{tschannen2024jetformer} and STARFlow~\cite{gu2025starflow} each design a new NF Transformer as the visual tower: JetFormer employs Jet~\cite{kolesnikov2024jet} to enable end-to-end continuous  AR modeling over raw image pixels, while STARFlow extends TARFlow~\cite{zhai2024normalizing} and demonstrates that continuous Autoregressive Flow can achieve competitive generation quality. 
But recent NF works~\cite{vnf,nice,nvp,glow,kolesnikov2024jet,zhai2024normalizing,gu2025starflow} predominantly map the data distribution to a standard Gaussian. This is a challenging objective, as forcing a high-dimensional and highly dispersed data distribution onto a simple isotropic Gaussian can introduce discontinuities or distortions, thus complicating the sampling process from the latent space and transforming back to the data space.

Inspired by the great work of Jetformer~\cite{tschannen2024jetformer}, we propose a framework named FARMER that leverages the strengths of both Normalizing Flows and Autoregressive models. 
As shown in \cref{fig:intro}, rather than mapping the data distribution to a fixed standard Gaussian, we employ an NF to transform  images into a latent sequence whose distribution is modeled implicitly by an AR model.
Concretely, we implement the NF with an Autoregressive Flow (AF) architecture, ensuring causal modeling for NF/AR within FARMER.
The two components are optimized jointly in an end-to-end fashion, preserving the tractable, exact likelihoods of NFs while endowing the target distribution with the expressivity of AR modeling.
Beyond this design, two inherent challenges remain:
(i) \textbf{Continuous AR over pixels}: Natural images are highly redundant. Without compression via VAEs~\cite{kingma2013auto,ldm} or discrete tokenizers~\cite{van2017neural,vqvae2}, directly modeling all pixels forces the AR model to handle extremely long-range pixel dependencies, and thus results in unstable training and sample quality degrading.
(ii) \textbf{Slow reverse inference in AF}: While AF substantially enhances the mapping capability via next-token modeling, they incur slow inference because the reverse inference process is strictly sequential.

To mitigate the redundancy in pixel AR modeling, we introduce a self-supervised dimension reduction mechanism that partitions NF latent channels into informative and redundant groups without information loss. 
The key insight is to factorize the token likelihood $P(z \mid c)$ as
\[
P(z \mid c)
\;=\;
P(z^{R} \mid z^{I}, c)\; P(z^{I} \mid c)
\;=\;
\Bigl[\prod_{i=1}^{N} P_{N+1}\!\bigl(z^{R}_{i} \mid z^{I}, c\bigr)\Bigr]
\;\Bigl[\prod_{i=1}^{N} P_{i}\!\bigl(z^{I}_{i} \mid z^{I}_{<i}, c\bigr)\Bigr],
\]
where $z^{I}$ denotes the informative channels and $z^{R}$ the redundant channels of each token. 
Concretely, the informative channels $z^{I}_{i}$ are modeled in the standard autoregressive manner, \ie, conditioned on the preceding informative tokens $z^{u}_{<i}$ and context $c$. 
The redundant channels $z^{R}_{i}$ across all tokens are modeled jointly by a shared distribution conditioned on the entire sequence of informative channels $z^{I}$ and context $c$.
This construction allows us to treat the redundant channels of all tokens as a single additional token, effectively converting $N$ high-dimensional tokens into $N\!+\!1$ lower-dimensional tokens. 
Maximizing the resulting token likelihood encourages FARMER to disentangle information across channel groups, 
\ie, concentrating contour and structural features in $z^{I}$, while assigning detail and color information to $z^{R}$, as illustrated in \cref{fig:dim_ablation}.

For the slow reverse issue of AF, we propose a one-step distillation scheme for efficient inference, which distills a single-step student reverse path from the  teacher's forward path, thereby avoiding the causal reverse process of AF models. 
Finally, we present a resampling-based Classifier-Free Guidance (CFG) algorithm 
that significantly improves generation quality in this framework.
In summary, we summarize our contributions as follows: 
\begin{itemize}
    \item We introduce FARMER, an elegant and powerful framework that jointly optimizes Autoregressive Flow and Autoregressive Transformer for continuous image pixel  likelihood estimation.
    \item We propose a self-supervised dimension reduction approach that simplifies modeling of high-dimensional visual data.
    \item We develop a one-step distillation method that accelerates AF reverse process by a factor of $22\times$ with only 60 additional training epochs, while maintaining comparable generation quality.
    \item We introduce a novel resampling-based CFG algorithm that substantially enhances generation quality.
\end{itemize}

\section{Preliminary}
\label{sec:pre}
\subsection{Normalizing Flows}
\label{subsec:nf}
Normalizing Flow~\cite{vnf,nice,nvp,glow,van2016pixel,iaf,maf,t_naf,kolesnikov2024jet,zhai2024normalizing} maps a complex data distribution $x \sim p_{data}(x)$ into a simple one $z \sim p_Z(z)$. The target distribution $p_Z(z)$ is usually chosen as a standard Gaussian, which is easy for density estimation and sampling. This transformation is achieved by applying a sequence of invertible functions $F={f_n\circ f_{n-1}\circ\dots\circ f_1}$.
Accordingly, the forward and inverse mappings are:
\begin{equation}
    z = F(x) = f_n \circ f_{n-1} \circ \dots \circ f_1(x),\qquad
    x = F^{-1}(z) = f_1^{-1} \circ f_2^{-1} \circ \dots \circ f_n^{-1}(z).
\end{equation}
Using the change-of-variables formula, NFs can calculate the exact probability density of a data point $x$ as:
\begin{equation}
p_{data}(x) = p_Z(z)\left| \det \left( \frac{\partial z}{\partial x} \right) \right|=p_Z(F(x)) \left| \det \left( \frac{\partial F(x)}{\partial x} \right) \right|,
\end{equation}
where $\det \left( \frac{\partial F(x)}{\partial x} \right)$ denotes the determinant of the Jacobian matrix of the transformation $F$. To facilitate training via maximum likelihood estimation, the learning objective is commonly formulated in terms of Negative Log-Likelihood (NLL):
\begin{equation}
\min_{F}\ -\log p_Z(F(x)) - \log \left| \det \left( \frac{\partial F(x)}{\partial x} \right) \right|.
\label{ep:4}
\end{equation}
Previous works~\cite{gu2025starflow,zhai2024normalizing} consider $p_Z$ as the standard Gaussian distribution $\mathcal{N}(0,1)$, so \eqref{ep:4} can be written as:
\begin{equation}
\label{eq:gaussian}
\min_{F}\ 0.5\left|| F(x)\right||_2^2 - \log \left| \det \left( \frac{\partial F(x)}{\partial x} \right) \right|.
\end{equation}

\subsection{AutoRegressive Models}
\label{subsec:ar}
AutoRegressive models formulate the likelihood of a token sequence $z = (z_1, z_2, \ldots, z_N)$ by factorizing it into a product of next-token conditional probabilities:
\begin{equation}
p(z)=\prod_{i=1}^{N} p(z_i | z_{<i}),
\label{eq:next-token}
\end{equation}
where $z_{<i}=(z_1,\ldots,z_{i-1})$  conditions only on the previous tokens $(z_1,\ldots,z_{i-1})$ to predict the next token.
Such AR paradigm has achieved remarkable scalability and tremendous success in language models~\cite{gpt4,gemini,qwen,llama,gemma}. Furthermore, it has also demonstrated promising capabilities in visual generation~\cite{llamagen,var,mar,mars,unified}. 

\section{Approach}
\subsection{Mapping Image to AR Distributions via Invertible Flows}
As aforementioned in \cref{eq:gaussian}, mapping high-dimensional and highly dispersed image data distribution to a simple isotropic Gaussian distribution via an NF can induce out-of-distribution issues and degrade the sampling quality~\cite{gu2025starflow}. Inspired by JetFormer~\cite{tschannen2024jetformer}, we propose a framework that combines the strengths of NF and AR models. Rather than using a fixed standard normal Gaussian, we employ an NF to transform images into a latent sequence whose distribution is modeled implicitly by an AR model. Then the NF and AR components are optimized jointly in an end-to-end fashion, preserving the tractable, exact likelihoods of NFs while endowing the target distribution with the expressivity of AR modeling.
The overall objective is to maximize the log-likelihood of the  via the change-of-variables formula: 
\begin{equation}
    \log p_{data}(x)=\sum^N_{i=1}\log p(z_i|z_{<i})+\log \left|\det(\frac{\partial F(x)}{\partial x})\right|,
\label{eq:total_loss1}
\end{equation}
where $z = F(x)$ denotes the forward mapping of the NF. The target distribution over $z$ is parameterized autoregressively.
To enhance the expressivity of the AR base, following JetFormer and GIVT~\cite{tschannen2023givt}, we model each conditional probability $p(z_i|z_{<i})$ with a Gaussian mixture model (GMM). The conditional log-likelihood for each token $z_i$ is:
\begin{equation}
    \log p(z_i|z_{<i})=\log(\sum^K_{k=1}\pi_{i,k}\mathcal{N}(z_i;\mu_{i,k},\sigma_{i,k}^2)),
\end{equation}
where the mixture weights $\pi_{i,k}$, means $\mu_{i,k}$, and deviations $\sigma^2_{i,k}$ are predicted by the AR model conditioned on preceding tokens $z_{<i}$.
Furthermore, different from Jetformer, we implement the NF model as an Autoregressive Flow (AF)~\cite{iaf,maf}. 
AF is a powerful universal approximator for distributions that adopt an autoregressive structure: the transformation of each token $z_i$ is conditioned only on the preceding tokens $z_{<i}$. 
Such AF architecture ensures that the entire pipeline maintains a consistent and powerful causal formulation. 
Notably, when the number of mixture components in GMM is set to one ($K=1$), the entire network, which composes an AF with an AR model, reduces to a single and deeper Autoregressive Flow. We provide a formal proof of this equivalence in \cref{app:k=1}.

\begin{figure}
    \centering
    \includegraphics[width=\linewidth]{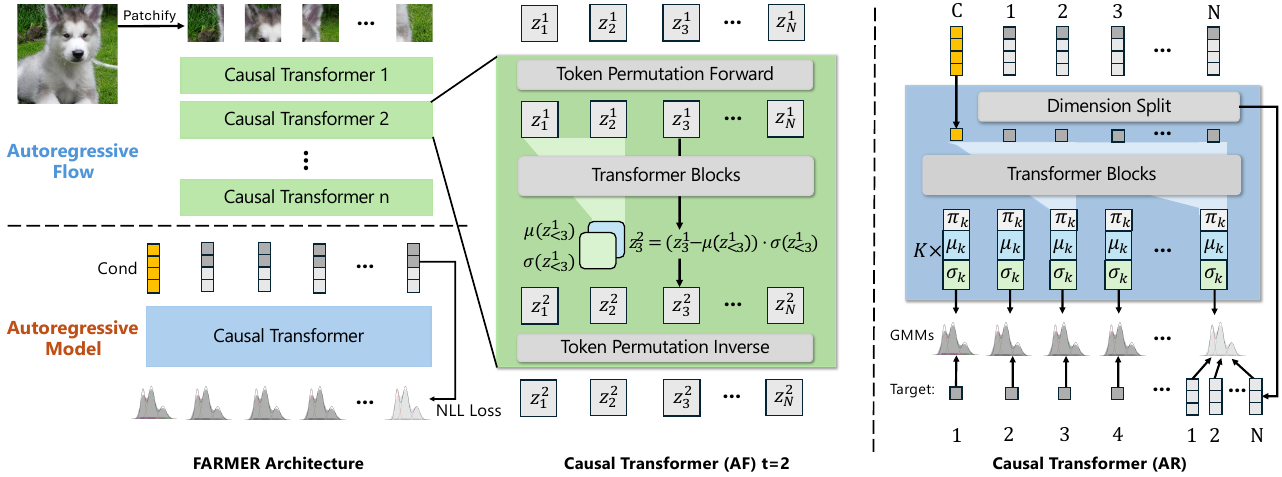}
    \caption{\textbf{Overview of FARMER.} \textbf{Left}, FARMER consists an autoregressive flow (AF) and an autoregressive (AR) model. The AF maps image patches to latent sequences, while the AR predicts Gaussian Mixture Models (GMMs) conditioned on these latents, optimizing their likelihood end-to-end.  \textbf{Middle}, Each AF block performs an invertible next-token transformation of the input sequence to obtain a new sequence. \textbf{Right}, AR splits latent channels into informative and redundant groups, modeling each informative token’s likelihood via a GMM conditioned on its previous tokens, and redundant tokens jointly via a shared GMM conditioned on all informative tokens. This separation enables disentangling structural and detailed information.}
    \label{fig:overview}
\end{figure}
\subsection{Flow AutoRegressive Transformer}
\label{sec:farmer}
We devise Flow AutoRegressive transforMER models (FARMER) that unify an invertible autoregressive flow with an autoregressive model into a single framework, which enables end-to-end training on raw image pixels by mapping the data onto an implicit  target distribution modeled by the AR. 

\textbf{Dequantize and Patchify}.
Specifically, given an input image $I \in \mathbb{R}^{H \times W \times C}$, FARMER first adds Gaussian noise to $I$.
It is a common practice~\cite{zhai2024normalizing,tschannen2024jetformer,cam} to add a small amount of noise to raw image $I$ to dequantize the discrete pixel values and create a more continuous data distribution. 
Following Jetformer~\cite{tschannen2024jetformer}, we enhance this technique by employing a noise augmentation strategy with annealed  noise levels. During training, we add Gaussian noise with a standard deviation $\mathcal{N}(0, \sigma^2)$ to $I$, where the noise level $\sigma$ is annealed from $0.1$ to $0.005$ via a cosine decay schedule.  
Then we patchify the noised image with a downsampling factor $p$ to obtain the patch representation $I' \in \mathbb{R}^{h \times w \times d}$, where $h = H/p$, $w = W/p$, and $d = C \cdot p^2$. 
Finally, we reshape $I'$ into a sequence of $N = h \cdot w$ continuous-valued visual tokens $X=\{x_1, x_2, \ldots, x_N\}$, with each token $x_i \in \mathbb{R}^d$. Notably, there is no dimension compression in the whole patchify process.

\textbf{Forward and Reverse of Autoregressive Flow}.
During training, FARMER utilizes an autoregressive flow $F$ to map token sequence $X\in \mathbb{R}^{N \times d}$ to latents $Z \in \mathbb{R}^{N \times d}$, \ie, $Z=F(X)$.
By design, $F$ is invertible (see \cref{fig:overview}) and composed of $n$ invertible blocks: $F=f_n \circ f_{n-1} \circ \dots \circ f_1$.
Letting $Z^0=X$ and $Z^n=Z$, the forward transformation for the $t$-th AF block, $Z^t=f_t(Z^{t-1})$, is defined for each token $z^{t}_{i}$ \footnote{Subscripts denote indexing $i$ along the token sequence dimension, and superscripts denote the $t$-th  AF block indices.} as follows:
\begin{equation}
   z^{t}_{i} = 
\begin{cases} 
z^{t-1}_{1} & \text{if } i=1, \\
\left( z^{t-1}_{i} - \mu_t(z^{t-1}_{<i}) \right) \odot \sigma_t(z^{t-1}_{<i}) & \text{if } i > 1, 
\end{cases}
\label{eq:maf_forward}
\end{equation}
where $z^{t-1}_{<i}$ represents the preceding tokens $\{z^{t-1}_{1},...,z^{t-1}_{i-1}\}$. The bias factor $\mu_t(z^{t-1}_{<i})$ and the scaling factor $\sigma_t(z^{t-1}_{<i})$ are predicted by $t$-th block conditioned on the preceding tokens $z^{t-1}_{<i}$ in a causal manner. 
Accordingly, the inverse of $t$-th block, $f_t^{-1}$, can be derived by algebraically solving for $z^{t-1}$ from the \cref{eq:maf_forward}. For each token, the inverse transformation $Z^{t-1}=f_t^{-1}(Z^t)$ is defined as (see \cref{fig:distll_a}): 
\begin{equation}
    z^{t-1}_{i} = 
\begin{cases} 
z^{t}_{1} & \text{if } i=1, \\
\left( z^{t}_{i} \oslash \sigma_t(z^{t-1}_{<i}) \right) + \mu_t(z^{t-1}_{<i}) & \text{if } i > 1, 
\end{cases}
\label{eq:maf_reverse}
\end{equation}
where $\oslash$ denotes element-wise division. 
For training via the change-of-variables formula, it is essential that the Jacobian determinant of each block $f_t$ can be efficiently computable.
Such autoregressive flow architecture enables that
the Jacobian  $\frac{\partial Z^t}{\partial Z^{t-1}}$  is lower triangular, so its determinant equals the product of its diagonal terms (\ie, the scaling factor $\sigma_t$). 
Consequently, the block-wise log-determinant is:
\[
\log \Bigl|\det \frac{\partial Z^{t}}{\partial Z^{t-1}}\Bigr|
= \sum_{i=1}^{N}\sum_{j=1}^{d} \log \bigl|[\sigma_t(z^{t-1}_{<i})]_j\bigr|
\]
By the chain rule, the total log-det of $F$ is the sum over blocks, which in our case reduces to:
\begin{equation}
\log \Bigl|\det \frac{\partial Z}{\partial X}\Bigr|
= \sum_{t=1}^{n} \log \Bigl|\det \frac{\partial Z^{t}}{\partial Z^{t-1}}\Bigr|
= \sum_{t=1}^{n}\sum_{i=1}^{N}\sum_{j=1}^{d} \log \bigl|[\sigma_t(z^{t-1}_{<i})]_j\bigr|.
\label{eq:detloss}
\end{equation}

\textbf{Permutation.}
To improve the expressiveness of AF, we follow TARFlow~\cite{zhai2024normalizing} and apply a permutation to the token sequence as shown in \cref{fig:overview}.
Specifically, at the beginning of the $t$-th AF block, we apply the forward permutation $\pi_t$ to $Z^{t-1}$, which reverses the token order.
After the forward AF transformation $Z^t=f_t(Z^{t-1})$, we apply the corresponding inverse permutation $\pi_t^{-1}$ to $Z^{t}$ to restore the original ordering.

\textbf{AR Modeling.}
After the AF forward mapping, we get the latent representation $Z=\{z_1,z_2,...,z_N\}$ from the input image.
Then we model its probability distribution with a large causal AR Transformer. 
The AR Transformer is conditioned on an embedding $c \in \mathbb{R}^{1\times D}$ which encodes conditional information such as a class label. 
To amplify its effect, we replicate condition embedding for $M$ times and prepend to the latent sequence $Z$. 
By the chain rule,
\[
P(Z|c) = \prod_{i=1}^{N} p(z_i | z_{<i}, c).
\]
For each token, the AR Transformer predicts the parameters of a $K$-component Gaussian Mixture Model (GMM) distribution $G_i$:
\begin{equation}
    p(z_i | z_{<i}, c) = \sum_{k=1}^{K} \pi_k(z_{<i}, c) \, \mathcal{N}\big(z_i \,;\, \mu_k(z_{<i}, c), \text{diag}(\sigma_k^2(z_{<i}, c))\big),
\label{eq:gmm}
\end{equation}
where $\pi_k\in \mathbb{R},\mu_k \in \mathbb{R}^d, \sigma_k \in \mathbb{R}^d$ are the mixture weights, means, and standard deviations of the $k$-th GMM component. 
To highlight the conceptual link to the invertible flow, \cref{eq:gmm} can be reformulated as:
\begin{equation}
    p(z_i | z_{<i}, c) = \sum_{k=1}^{K} \pi_k(z_{<i}, c) \, \mathcal{N}\big((z_i-\, \mu_k(z_{<i},c))\odot\text{diag}(\frac{1}{\sigma_k(z_{<i}, c)})\,;\, \mathbf{0}, I_d\big) \left|\frac{1}{\sigma_k(z_{<i}, c)}\right|,
\label{eq:gmm2}
\end{equation}
This formulation reveals that each GMM component models $z_i$ by a simple and invertible affine transformation (shifting by $\mu_k$ and scaling by $\frac{1}{\sigma_k}$) to a random variable drawn from a standard Gaussian distribution.
This reveals that each GMM component performs an invertible affine normalization, \ie, 
$ \frac{(z_i - \mu_k)}{\sigma_k} \sim \mathcal{N}(0, I)$.

\textbf{Learning Objective.}
As described in \cref{eq:total_loss1}, \cref{eq:detloss} and \cref{eq:gmm}, the training loss of FARMER is the negative log-likelihood (NLL) of data and averaged over all dimensions:
\begin{equation}
\mathcal{L} = -\frac{1}{N \cdot d} \left( \sum_{i=1}^{N} \log p(z_i | z_{<i}, c) + \log \left| \det \frac{\partial Z}{\partial X} \right| \right).
\label{eq:total_loss2}
\end{equation}

\subsection{Self-supervised Dimension Reduction}
A fundamental challenge in pixel AR modeling is redundancy:
natural images are intrinsically low-dimensional signals whose spectrum are dominated by low frequencies~\cite{tschannen2024jetformer}. 
Although an invertible AF can faithfully map the data distribution, its bijective nature preserves dimensionality.
For a $256 \times 256 \times 3$ image with patch size $16$, the latent sequence has $N = (256/16)^2 = 256$ tokens, each of dimension $d = 768$. 
This high-dimensional latent $Z$ exacerbates two issues:
(i) per-token AR modeling with a K-component GMM in $\mathbb{R}^d$ becomes exceptionally challenging.
(ii) The enlarged latent volume expands the sampling space, reducing efficiency and often degrading sample quality.

Prior work like RealNVP~\cite{nvp} factors out half of the dimensions and model them with Gaussian priors. 
Jetformer~\cite{tschannen2024jetformer} follows a similar strategy: it models the informative dimensions $Z^{I}$ autoregressively and assigns the redundant dimensions $Z^{R}$ a standard Gaussian prior, effectively assuming
\[
P(Z \mid c) = P(Z^{R}) \, P(Z^{I} \mid c),
\]
i.e., $Z^{R}$ is independent of both $Z^{I}$ and $c$.
This is a strong assumption that is often violated in practice: informative and redundant parts typically remain correlated, so enforcing independence can discard information.
Moreover, decoupling $Z^{R}$ from $c$ and $Z^{I}$ restricts how other modalities interact with the full latent, leading to suboptimal performance on multi-modal tasks.

To this end, we propose a novel self-supervised dimension reduction technique to address the above issues. 
It reduces the complexity of AR modeling, shrinks the  sampling space, and lowers the computational cost, all while avoiding information loss. 
As shown in \cref{fig:overview}
we split the latent $Z \in \mathbb{R}^{N \times d}$ channel-wise  into an informative part $Z^{I} \in \mathbb{R}^{N \times d^{I}}$ and a redundant part $Z^{R} \in \mathbb{R}^{N \times d^{R}}$, with $d = d^{I} + d^{R}$.
Then we correctly factorize the joint probability via the chain rule:
\[
P(Z \mid c) = P(Z^{I} \mid c)\, P(Z^{R} \mid Z^{I}, c).
\]
Rather than assuming that $Z^{R}$ is independent of ($Z^{I},c$) in Jetformer, we explicitly condition $Z^R$ on both $c$ and $Z^I$, where $Z^I$ serves as the global image context. 
Furthermore, we constrain all tokens in $Z^{R}$ to share a GMM distribution, while modeling tokens in $Z^{I}$ in a token-by-token manner. 
This design encourages self-supervised disentanglement of distinct information across channel groups, without relying on a standard Gaussian prior.

For $P(Z^{I}|c)$, we model each informative token $z^{I}_i$ autoregressively with an individual GMM distribution $G_i$ predicted by the AR Transformer conditioned on $c$ and the preceding $z^{I}_{<i}$, thereby being capable of capturing complex distributions. 
In contrast, for $P(Z^{R}|Z^{I},c)$, 
we use the entire informative sequence $Z^{I}$ (global context) together with $c$ to predict a single shared GMM $G_{N+1}$ for all redundant tokens $z^{R}_i$,
By maximizing the combined likelihoods, our method successfully encourages complex contour and structural information to be reserved into $Z^{I}$, while the simple color and fine-detail information is relegated to $Z^{R}$, as shown in \cref{fig:dim_ablation} and discussed in \cref{sec:dim_seperate}.

After dimension reduction, the final training loss $\mathcal{L}$ is rewritten as the sum of the NLL for both components:
\begin{equation}
\mathcal{L} = -\frac{1}{N \cdot D} \left( \sum_{i=1}^{N} \log p(z^{I}_i | z^{I}_{<i}, c) + \sum_{i=1}^{N} \log p(z^R_i | z^I_{\leq N}, c) + \log \left| \det \frac{\partial Z}{\partial X} \right|\right).
\label{eq:total_loss3}
\end{equation}

\subsection{Resampling-based Classifier-Free Guidance}
\label{sec:cfg}
Classifier-Free Guidance (CFG) has become a standard technique for improving sample quality in diffusion~\cite{ldm,dit,sit} and autoregressive models~\cite{llamagen,var,mar}.
Conceptually, CFG steers the sampling process from a base distribution towards a target conditional distribution. 
For FARMER, the guided log-probability for a latent token $z$ can be formulated as:
\begin{equation}
    \log p'(z) \propto \textcolor{blue}{\log p_c(z)} + \textcolor{red}{w \cdot (\log p_c(z) - \log p_u(z))} = \textcolor{blue}{\log p_u(z)} + \textcolor{red}{(w+1) \cdot (\log p_c(z) - \log p_u(z))},
    \label{eq:cfg}
\end{equation}
where $p_c(z)=p(z|c)$ is the conditional GMM distribution, $p_u(z)=p(z|\emptyset)$ is the unconditional GMM, and $w$ is the guidance scale. 
However, the guided distribution $p'(z)$ is a product and sum of GMMs, which does not correspond to any known tractable distribution, making direct sampling infeasible.

To make it practical, we introduce a novel Resampling-based CFG. 
The key insight is that the target distribution $p'(z)$ can be decomposed into two components as shown in \cref{eq:cfg}: the first term (\textcolor{blue}{blue}) is a tractable GMM distribution and can be sampled directly, while the second term (\textcolor{red}{red}) is not  samplable
but allows evaluation of the sample probability under such distribution. 
Therefore, we approximate the sampling from $p'(z)$ via a three-step resampling scheme as detailed in \cref{alg:cfg_sampling}. For For each token $z_i$, the procedure is: 
(i) \textbf{Propose}. Sample $s$ candidates from the conditional GMM $p_c(z_i)$ and $s'$ candidates from the unconditional GMM $p_u(z_i)$ respectively.
(ii) \textbf{Weigh}. Compute the corresponding log probability of each candidates as the second term in \cref{eq:cfg}, and normalize these weights. 
(iii) \textbf{Resample}. Resample from the categorical distribution that consists of the normalized weights of all candidates, to obtain the final sample. 
In summary, the probability where candidate $z$ is selected in the ``propose'' step is $p_c(z)$, and that in the ``resample'' step is $\left(\frac{p_c(z)}{p_u(z)}\right)^w$. 
This resampling procedure ensures that the overall probability $p_c(z)\left(\frac{p_c(z)}{p_u(z)}\right)^w$ matches the target probability $p'(z)$. More details are provided in the \cref{app:cfg}.

\begin{figure}[t]
    \centering
    \begin{subfigure}[t]{0.57\linewidth}
        \centering
        \includegraphics[height=5.0cm]{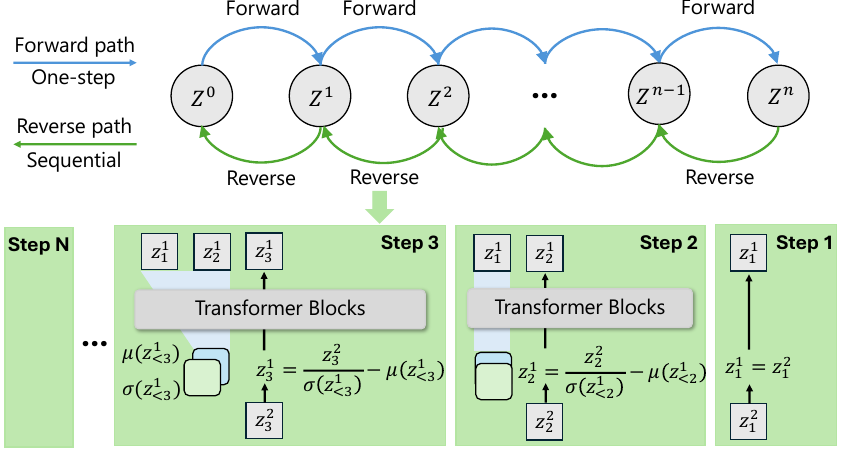}
        \caption{Autoregressive Flow Reverse Process}
        \label{fig:distll_a}
    \end{subfigure}
    \begin{subfigure}[t]{0.42\linewidth}
        \centering
        \includegraphics[height=5.0cm]{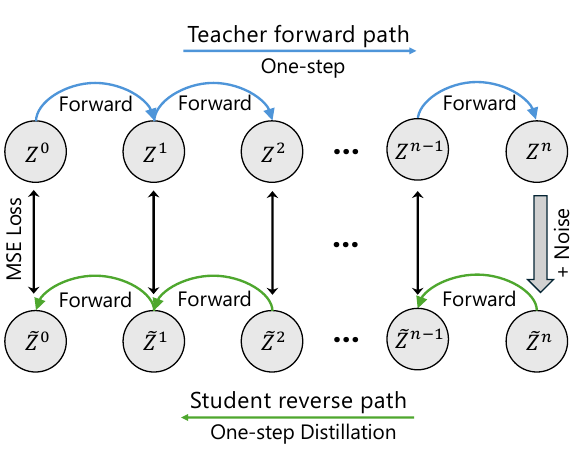}
        \caption{One-Step Distillation Process}
        \label{fig:distll_b}
    \end{subfigure}
    \caption{\textbf{One-Step Distillation.} (a) The autoregressive flow (AF) reverse process reconstructs tokens sequentially, conditioning each token on previous ones, which leads to slow inference. (b) Our method distills a one-step student reverse path from the frozen teacher forward path in an end-to-end manner, approximating the reverse process of each AF block by the corresponding student AF block’s forward process, thereby  enabling  $22\times$ faster AF reverse process  and $4\times$ overall
 inference speed-up.}
    \label{fig:distll}
\end{figure}

\subsection{Fast Inferring via One-Step Distillation}
A significant drawback of the Autoregressive Flow is the slow inference speed, caused by its sequential and token-by-token reverse process. 
As shown in \cref{eq:maf_reverse}, during the inverse mapping ($f^{-1}_t$) of AF block $t$, the calculation of each token $z_{t-1,i}$ is conditioned on the preceding tokens $z_{t-1,<i}$, leading to a complexity of $\mathcal{O}(N\times n)$. 
Such autoregressive dependency brings a substantial inference speed bottleneck, and such limitation is also noted in recent AF-Transformer works like TARFlow~\cite{zhai2024normalizing} and STARFlow~\cite{gu2025starflow} whose token sequence length is 1024 with the patchsize of 8.   

Beneficial from the invertible nature of Normalizing Flow, whose forward and reverse paths are exact inverses, we can train a new AF whose forward path 
mirrors the original AF’s reverse path.
Furthermore, because the forward/reverse path of NF consist of finite steps, we can invert the original AF’s  forward path $(Z_0,Z_1,...,Z_n)$  to obtain its reverse path $(Z_n,Z_{n-1},...,Z_0)$, and utilize such reverse path to supervise the new AF,
thereby avoiding the original AF to perform slow reverse process to obtain its reverse path.

As shown in \cref{fig:distll} and inspired by the generative distillation works~\cite{salimans2022progressive,yin2024one,walton2025distilling}, we propose a one-step distillation scheme that learns a single-step student reverse path from the trained teacher’s forward path while maintaining competitive sample quality.
\cref{alg:farmer_distill} details the procedure: we first obtain a teacher AF model, trained within the FARMER framework.
Then, we  initialize the student by copying the teacher AF and enable its attention bidirectional. 
At each distill iteration, we forward training data $z_0$  to the latent $z_n$ by the teacher AF. 
In this way, we obtain a teacher forward path $F(Z^0)=(Z^0,Z^1,...,Z^n)$.
We use its reversal $(Z^n,Z^{n-1},...,Z^0)$ as the supervision target for the student’s forward path $G(\tilde{Z}^n)=(\tilde{Z}^n,\tilde{Z}^{n-1},...,\tilde{Z}^0)$. 
Specifically, to enhance the robustness of the student AF, we add a small Gaussian noise to $Z^n$ as $\tilde{Z}$ and take $\tilde{Z}$ as the input of the student AF. 
Then, the output latent $\tilde{Z}^{t-1}$ of each $t$-th student AF block is supervised by minimizing the Mean Squared Error (MSE) against the $Z^{t-1}$ from the teacher path. By distilling one such student AF model, we significantly accelerate the reverse process from 0.1689 to 0.0076 seconds per image. 
As discussed in \cref{sec:exp_analysis} and \cref{tab:inference_accelerate}, such one-step distillation
brings a $22\times$ acceleration for AF reverse process while maintaining comparable generation quality.

Notably, different from the progressive distillation for diffusion models, our approach offers three main advantages: it distills the entire AF model in an end-to-end manner, ensuring robustness to accumulative inference error; it eliminates the need for teacher models to run the inference process, thereby accelerating the distillation process; and it requires only 60 additional training epochs on the AF.

\begin{minipage}{0.46\textwidth}
\begin{algorithm}[H]
\centering
\caption{Resampling-based CFG method}\label{alg:cfg_sampling}
\begin{algorithmic}
\Require AR model $P_\theta$ and AF model $F_\theta$
\Require Guidance scale $w$
\For{{\setlength{\fboxsep}{0pt}{$i\in[0,...,N+1]$}}} \Comment{Sampel tokens}
\State {\setlength{\fboxsep}{0pt}{\texttt{\# step1:Propose candidates}}}
\State $G_{c,i}=P_\theta(z^u_{<i}\,;c)$ \Comment{Predict GMM$_c$}
\State $G_{u,i}=P_\theta(z^u_{<i}\,;\emptyset)$ \Comment{Predict GMM$_u$}
\State $z_{i,j} \sim G_{c,i},\ j\in[0,..,s]$ \Comment{Sample from $p_c(z)$}
\State $z_{i,j} \sim G_{u,i},\ j\in[{s+1},..,s+s']$ \Comment{from $p_u(z)$}
\State {\setlength{\fboxsep}{0pt}{\texttt{\# step2:Weigh candidates}}}
\If{$j\in [0,...,s]$} \Comment{Calculate weights}
\State $\pi_j=w\cdot(\log G_{c,i}(z_{i,j})-\log G_{u,i}(z_{i,j}))$
\Else
\State $\pi_j=(w+1)\cdot (\log G_{c,i}(z_{i,j})-\log G_{u,i}(z_{i,j}))$
\EndIf
\State $\pi_1,...,\pi_{s+s'}=\textbf{logsoftmax}(\pi_1,...,\pi_{s+s'})$
\State {\setlength{\fboxsep}{0pt}{\texttt{\# step3:Resample from candidates}}}
\If{$i \le N$} \Comment{For informative tokens}
\State $idx\sim \textbf{Categorical}(\pi_1,...,\pi_{s+s'})$
\State $z^u_i:=z_{i,idx}$
\Else \Comment{For redundant tokens}
\For{$k\in[0,...,N]$} \Comment{$s,s'$ larger in here}
\State $idx_k\sim \textbf{Categorical}(\pi_1,...,\pi_{s+s'})$ 
\State $z^d_k:=z_{idx_k}$ 
\EndFor
\EndIf

\EndFor
\State $z=\textbf{concat}[[z^u_1,z^d_i],...,[z^u_N,z^d_N]]$
\State $x=F^{-1}_{\theta}(z)$ \Comment{Reverse to data}
\end{algorithmic}
\end{algorithm}
\end{minipage}
\hfill
\begin{minipage}{0.49\textwidth}
\begin{algorithm}[H]
\centering
\caption{One-step sampling distillation}\label{alg:farmer_distill}
\begin{algorithmic}
\Require {\setlength{\fboxsep}{0pt}{Trained teacher AF (frozen) \\$\mathbf{F_\eta}={f_{\eta_n}\circ f_{\eta_{n-1}}\circ\dots\circ f_{\eta_1}}$}}
\Require Data set $\mathcal{D}$
\Require {\setlength{\fboxsep}{0pt}{Student AF $\mathbf{G_\theta}={g_{\theta_1}\circ g_{\theta_{2}}\circ\dots\circ g_{\theta_n}}$}}
\For{{\setlength{\fboxsep}{0pt}{$m$ epochs}}}
\For{{\setlength{\fboxsep}{0pt}{$K$ iterations}}}
\State $x \sim \mathcal{D}$
\State $x = \textbf{Patchify}(x)$ 
\State $Z^0:=x$ 
\For{{\setlength{\fboxsep}{0pt}{$n$ Teacher AF blocks}}}
\State $Z^t, \_ = f_{\eta_t}(Z^{t-1})$ \Comment{Teacher Transform}
\EndFor
\State $Z:=Z_n$
\State $\epsilon \sim N(0, I)$ \Comment{Sample noise}
\State $s \sim U[0,0.3]$ \Comment{Sample scale}
\State $\tilde{Z} = Z + s\cdot \epsilon$ \Comment{Add noise to latent}
\State $\tilde{Z}^n:=\tilde{Z}$
\State {\setlength{\fboxsep}{0pt}{\texttt{\# Distill a one-step student reverse}}}
\State {\setlength{\fboxsep}{0pt}{\texttt{\# path from the teacher forward path}}}
\For{{\setlength{\fboxsep}{0pt}{$n$ Student AF reversed blocks}}}
\State $\tilde{Z}^{t-1}, \_ = g_{\theta_t}(\tilde{Z}^{t})$ \Comment{Student Transform}
\State $L_{\theta_t}=\lVert \tilde{Z}^{t-1}-Z^{t-1}\rVert_{2}^{2}$ \Comment{MSE loss}
\EndFor
\State $L_{\theta} = \frac{1}{n}\sum_t^nL_{\theta_t}$
\State $\theta \leftarrow \theta - \gamma\nabla_{\theta}L_{\theta}$
\EndFor
\EndFor
\State \vspace{0.12cm}
\end{algorithmic}
\end{algorithm}
\end{minipage}

\section{Experiments}
\subsection{Experimental Settings}
\textbf{Datasets}. We empirically verify the merits of the proposed FARMER for image generation on ImageNet~\cite{imagenet} dataset at 256 $\times$ 256 resolution, which consits of 1,281,167 training images from 1K different classes. 

\textbf{Network Architectures.} We design two model scales: FARMER-1.1B/1.9B. The number of invertible AF blocks is set to 28 and 32 respectively. Each AF block contains 4/6 Transformer layers for FARMER-1.1B/1.9B. For the AR Transformer module, the number of Transformer blocks is 12 and 24 respectively.  
For the GMM prediction heads, the informative dimensions ($d^I$) are set to 128 with $K=64$ mixtures, while redundant dimensions ($d^R$) are set to 640 with $K=200$ mixtures. Table~\ref{tab:farmer_conf} summarizes the detailed architectural configurations.

\begin{table}[!th]\small
    \centering
    \caption{The architecture configurations of FARMER in two different scales (i.e., 1.1B and 1.9B). }
    \vspace{-0.5em}
    \setlength\tabcolsep{2.0pt}
    \begin{tabular}{l|ccc|cccc|c}
         \toprule
         \multirow{2}{*}{Model} & \multicolumn{3}{c|}{Autoregressive Transformer} & \multicolumn{4}{c|}{Invertible Autoregressive Flow} & \multirow{2}{*}{Params} \\
                       & ~Layers          & ~Hidden size   & ~Params       & ~~AF Blocks        & ~Layers        & ~Hidden size       & ~Params       &                           \\ 
        \toprule
        FARMER-1.1B & 12 & 768  & 295M & 28 & 4 &  768 & 828M & 1.1B \\
        FARMER-1.9B & 24 & 1024 & 498M & 32 & 6 & 768 & 1.4B & 1.9B \\ 
        \bottomrule
    \end{tabular}
    \label{tab:farmer_conf}
\end{table}

\textbf{Training Setup}. We train the models using AdamW optimizer ($\beta_1=0.9, \beta_2=0.95$) with weight decay of 0.03 for 320 epochs. A cosine learning rate schedule is applied, starting from $1\times10^{-4}$ to $1\times10^{-6}$, with 5,000-step linear warmup. Gaussian noise with a cosine decay from 0.1 to 0.005 is added to the raw image.  

\textbf{Evaluation Metrics.} To assess sample quality, we use Fr\'echet Inception Distance (FID) \cite{fid}, Inception Score (IS) \cite{is} and Precision/Recall \cite{precision} on 50K generated samples to measure the image quality on ImageNet-256.

\begin{table}[t]
\centering
\caption{
\textbf{System performance comparison} on ImageNet $256\times256$ class-conditioned generation. {\small ``$\downarrow$'' or ``$\uparrow$'' indicate lower or higher values are better. Metrics include Fréchet inception distance (FID), inception score (IS), precision and recall. \textbf{Resampling-based CFG} is applied on FARMER.}
}
\vspace{-0.5em}
\small 
\setlength{\tabcolsep}{1.8mm}{
\begin{tabular}{c|c|c|c|cccc}
\toprule
Types & Models & Params & ~Epochs~
& ~FID$\downarrow$ & IS$\uparrow$ & Pre.$\uparrow$ & Rec.$\uparrow$  \\
\midrule
\multicolumn{7}{l}{\textit{Latent Generative Models}} \\
\multirow{6}{*}{Diff.} & LDM-4~\cite{ldm} & 400M + 86M & 170 & 3.6 & 247.7 & {0.87} & 0.48 \\

&DiT-XL~\cite{dit} & 675M + 86M & 1400 & 2.27 & 278.2 & {0.83} & 0.57 \\

&SiT-XL~\cite{sit} & 675M + 86M & 1400 & 2.06 & 270.3 & 0.82 & 0.59 \\

&FlowDCN~\cite{flowdcn} & 618M + 86M & 400 & 2.00 & 263.1 & 0.82 & 0.58 \\

&REPA~\cite{repa} & 675M + 86M & 800 & 1.42 & 305.7 & 0.80 & 0.64 \\

&DDT-XL~\cite{decoupled_dit} & { 675M + 86M} & {400} & {1.26} & {310.6} & {0.79} & {0.65} \\
&REPA-E~\cite{leng2025repa} & { 675M + 86M} & {800} & {1.12} & {302.9} & {0.79} & {0.66} \\
\midrule
\multirow{3}{*}{AR} & GIVT~\cite{tschannen2023givt} & {1.67B+53M} & 500 & 2.59 &- & 0.81 & 0.57  \\
&MAR-AR~\cite{mar} & {479M+66M} & {800} & 4.69 & 244.6 & - & - \\
&{MAR-L~\cite{mar}} & {479M + 66M} & {800} & {1.78} & {296.0} & {0.81} & {0.60} \\
\midrule
\multirow{2}{*}{NF} & STARFlow~\cite{gu2025starflow} one-step denoise& {1.4B+86M} & 320 & 2.96 & - & - & - \\
& STARFlow~\cite{gu2025starflow} finetune decoder& {1.4B+86M} & 320 & 2.40 & - & - & - \\
\midrule
\multicolumn{7}{l}{\textit{Pixel Generative Models}} \\
\multirow{1}{*}{GAN} & BigGAN~\cite{biggan}  &112M & / & 6.95  & 224.5  & 0.89 & 0.38 \\
\midrule
\multirow{5}{*}{Diff.} & ADM~\cite{adm} & 554M & 400  & 4.59 & 186.7 & 0.82 & 0.52  \\
&CDM~\cite{cdm} & - & 2160 & 4.88 & 158.7 & - & -  \\
&SimpleDiffusion~\cite{simple} & 2.0B & 800 & 2.77 & 211.8 & - & -  \\
&PixelFlow-XL/4~\cite{chen2025pixelflow} & 677M & 320 & 1.98 & 282.1 &  0.81 & 0.60 \\
&PixNerd-XL/16~\cite{wang2025pixnerd} & 700M & 320 & 1.93 & 298 & 0.80 & 0.60 \\
&SiD2 patch 1~\cite{hoogeboom2024simpler} & - & 1280 & 1.38 & - & - & - \\ 
\midrule
\multirow{1}{*}{AR} &FractalMAR-H \cite{fractal} & 844M & 600 & 6.15 & 348.9 & 0.81 & 0.46 \\
\midrule
\multirow{2}{*}{NF} &TARFlow~\cite{zhai2024normalizing} patch 8 & 1.3B & 320 & 5.56 & - & - & - \\
 &STARFlow~\cite{gu2025starflow} patch 8 & 1.4B & 320 & 4.69 & - & - & - \\
\midrule
\rowcolors{2}{gray!20}{white}
\multirow{5}{*}{NF+AR} &JetFormer~\cite{tschannen2024jetformer} & 2.8B & 500 & 6.64 & - & 0.69 & 0.56 \\
 & \cellcolor[gray]{.92}FARMER 1.1B patch 16 & \cellcolor[gray]{.92}1.1B & \cellcolor[gray]{.92}320 & \cellcolor[gray]{.92}5.40 & \cellcolor[gray]{.92}212.23 & \cellcolor[gray]{.92}0.78 & \cellcolor[gray]{.92}0.45 \\
 & \cellcolor[gray]{.92}FARMER 1.1B patch 8 & \cellcolor[gray]{.92}1.1B & \cellcolor[gray]{.92}320 & \cellcolor[gray]{.92}5.02 & \cellcolor[gray]{.92}237.00 & \cellcolor[gray]{.92}0.80 & \cellcolor[gray]{.92}0.45 \\
 & \cellcolor[gray]{.92}FARMER 1.9B patch 16  & \cellcolor[gray]{.92}1.9B & \cellcolor[gray]{.92}320 & \cellcolor[gray]{.92}3.96 & \cellcolor[gray]{.92}250.64 & \cellcolor[gray]{.92}0.79 & \cellcolor[gray]{.92}0.50 \\
 & \cellcolor[gray]{.92}FARMER 1.9B patch 8  & \cellcolor[gray]{.92}1.9B & \cellcolor[gray]{.92}320 & \cellcolor[gray]{.92}3.60 & \cellcolor[gray]{.92}269.21 & \cellcolor[gray]{.92}0.81 & \cellcolor[gray]{.92}0.51 \\

\bottomrule
\end{tabular}}
\label{tab:imagenet256_sota}
\end{table}

\subsection{Results}
\textbf{System-level Comparison}.
As shown in \cref{tab:imagenet256_sota}, we compare FARMER with various generative models, including both latent and pixel-based approaches, on the class-conditional ImageNet 256$\times$256 benchmark. 
Notably, FARMER significantly outperforms JetFormer~\cite{tschannen2024jetformer}, the most comparable baseline to our model, reducing the FID by 3.04. Furthermore, FARMER demonstrates superior generation quality compared to the NF-based models, TARFlow~\cite{zhai2024normalizing} and STARFlow~\cite{gu2025starflow}. FARMER also achieves competitive performance and faster convergence speed against mainstream Generative Adversarial Networks (GANs), diffusion models, and AR models. While methods like PixelFlow~\cite{chen2025pixelflow} and PixNerd~\cite{wang2025pixnerd} employ complex multi-stage pipelines to achieve better results, our approach remains highly competitive by utilizing a simple, single-stage, end-to-end training strategy. Compared to latent generative models, our method maintains strong generative performance. Latent generative models often benefit from a well-structured continuous latent space, modeled by VAEs, that facilitates high-quality sampling. However, by operating directly in pixel space, our model gains direct access to the raw data distribution. This approach can potentially capture more detailed data semantics without the information bottleneck imposed by VAEs.

\begin{figure}[t]
    \centering
    \includegraphics[width=\linewidth]{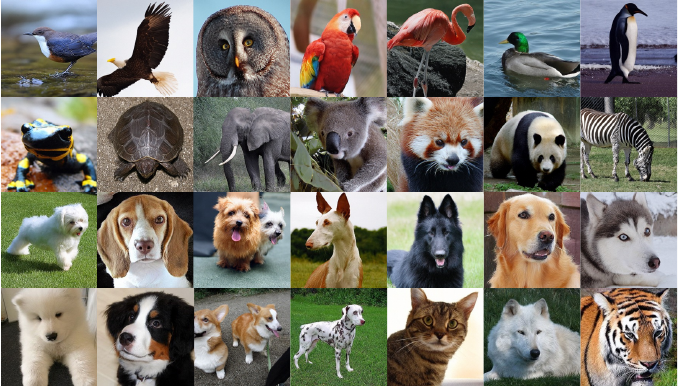}
    \vspace{-2.0em}
    \caption{\textbf{Qualitative Results}. Images generated by FARMER on ImageNet 256x256.}
    \label{fig:quality}
\end{figure}

\begin{figure}[t]
    \centering
    \includegraphics[width=\linewidth]{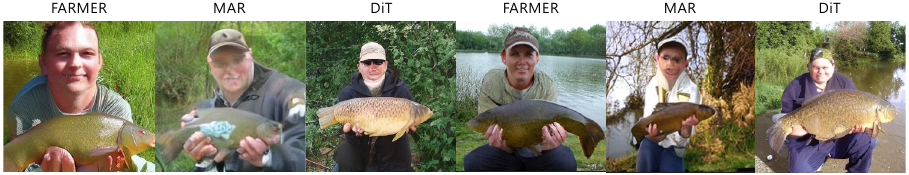}
    \vspace{-2.0em}
    \caption{\textbf{Qualitative Comparison}. Images of class 0 in ImageNet generated by FARMER, MAR, and DiT.}
    \label{fig:quality_b}
\end{figure}

\textbf{Qualitative Results}.
To qualitatively evaluate FARMER, we show 28 generated images by FARMER-1.9B in \cref{fig:quality}, sampling using resampling-based classifier-free guidance. As shown, our FARMER generates diverse images with high quality. A key advantage of FARMER over latent generative models is its ability to preserve fine-grained details. This is because our end-to-end training directly accesses the raw data distribution, and the invertible nature of NFs prevents information loss. As shown in \cref{fig:quality_b}, our FARMER can reconstruct intricate features, such as faces, which are often blurred or distorted by the compression of VAEs.

\subsection{Experimental Analysis}\label{sec:exp_analysis}
\textbf{Ablation Study}.
Here we investigate the impact of each component within the FARMER framework on overall performance. \cref{tab:abla} reports the performance of FARMER-1.1B with GMM components number (K=1024) across the ablated runs of different components on ImageNet 256$\times$256 dataset for class-conditional image generation. Natural images typically possess a high degree of redundancy, and low-dimensional signals with low-frequency components dominating the spectrum~\cite{tschannen2024jetformer}. Direct transformation of original images using normalizing flows yields latent representations with unchanged dimensionality. Partitioning these high-dimensional latents into equal-length, high-dimensional tokens complicates AR modeling and sampling. By introducing a self-supervised dimension reduction design as \cref{eq:total_loss3}, the FID notably decreases from 61.17 to 49.29, and IS also improves from 22.10 to 30.61. Next, we repeat the class embedding 64 times to enhance the conditional guidance, the FID further decreases to 45.34. If we consider the AR model as a block of AF, adding a token permutation operation between AF and AR is beneficial to preserve the fixed dependency between token sequences. The FID further decreases to 44.56. CFG is essential for improving generation quality in modern generative models during sampling. We first adopt a naive CFG sampling method from Jetformer~\cite{tschannen2024jetformer}, the FID score notably decreases to 8.66. Then, we upgrade the CFG sampling method to the resampling-based method described in \cref{sec:cfg}, the FID score further decreases to 5.67. Together, these design choices enable FARMER to achieve strong performance across most evaluation metrics.
\begin{table}[]
\centering
\caption{
\textbf{Ablation study of FARMER.} We demonstrate relative impact of various components on generation quality.
}\label{tab:abla}
\vspace{-0.8em}
\scalebox{1}
{\begin{tabular}{cccc|cc}
\toprule
    Self-supervised Dim. Reduce & Condition Repeat  & Final Permute  & CFG Method &FID$\downarrow$ & IS$\uparrow$ \\
\midrule
  \cha & \cha & \cha & \cha & 61.17 & 22.10 \\
\midrule
 \gou  & \cha & \cha  & \cha & 49.29 & 30.61  \\
 \gou  & \gou  & \cha  & \cha & 45.34 & 33.87 \\
 \gou  & \cha   & \gou  & \cha & 45.69 & 33.73  \\
 \gou  & \gou   & \gou   & \cha & 44.56  & 33.17  \\
 \gou  & \gou   & \gou   & Naive Method & 8.66 & 233.84 \\
 \gou  & \gou   & \gou  & Resampling-based & 5.67 & 215.53 \\
\bottomrule
\end{tabular}}
\end{table}

\begin{table}[]
\caption{\textbf{Impact of Normalizing Flow Architectures}.}
\vspace{-1.0em}
    \centering
    \begin{tabular}{l|cccc}
    \toprule
      NF Architectures& FID & IS & Forward Speed (s/img) & Reverse Speed (s/img) \\
       \midrule
                  Jet&106.23& 13.14&   0.0065      &     0.0099    \\
                  AF & 5.55& 194.63&   0.0066      &     0.1689    \\
 AF+One-step Distll. & 5.63& 193.49&   0.0066      &     0.0076    \\
    \bottomrule
    \end{tabular}
    \label{tab:arch}
\end{table}

\textbf{Impact of Normalizing Flow Architectures}.
The architecture design of NF is an important research topic and has been extensively studied~\cite{vnf,nice,nvp,glow,van2016pixel,iaf,maf,kolesnikov2024jet,zhai2024normalizing}. Different NF architectures exhibit distinct characteristics in terms of representational capacity, training speed, and inference efficiency. Here we primarily compare two architectures, Jet and AF, which have demonstrated strong performance in modern generative models Jetformer~\cite{tschannen2024jetformer} and Tarflow~\cite{tschannen2024jetformer}, respectively. For a fair comparison, we employ similar network parameters, same block numbers, the same layers per block, and the same AR models. Their representational capacity is evaluated using the FID metric, while forward and reverse speeds are also reported. \cref{tab:arch} summarizes these results. Specifically, in each transformation of Jet, Jet first computes an affine transform from one half of the input latent channels by a Jet block and then applies it to the other half of the input channels; this pattern applies to both forward and reverse passes. Jet is constructed by stacking N such transformations. This simple and efficient design enables Jet to achieve fast forward and reverse computations, but it also limits its representational capacity, leading to a failure to separate different information of the image into two channel groups. As described in \cref{sec:farmer}, in each transformation of AFs, each token is updated based on preceding tokens through the block, resulting in a reverse process where each token must be generated one by one. This enhances representation ability but leads to slow reverse speed. To address this, we introduce a one-step distillation strategy. By distilling a student AF model from the trained and frozen teacher AF model over only 60 additional training epochs on the NF, we significantly improve the reverse speed from 0.1689 seconds per image to 0.0076 seconds per image. This approach provides a fast and expressive architecture for both training and inference.

\textbf{Dimension Reduction Method Comparison}.
We also compare our self-supervised dimension reduction method with the approach adopted in JetFormer~\cite{tschannen2024jetformer}. JetFormer assumes that a subset of channels is redundant and independent from the remaining channels and maximizes the likelihood of these redundant channels under the standard Gaussian prior. This assumption may result in information loss, thereby degrading generation quality. In contrast, our self-supervised method models redundant channels as being conditionally dependent on informative channels which encapsulate the global information of images. Our method achieves improved generative performance, reducing FID from 7.81 to 5.67, and increasing IS from 182.87 to 215.53.

\begin{figure}[htbp]
    \centering
    \begin{subfigure}[]{0.46\linewidth}
        \centering
        \includegraphics[width=\linewidth]{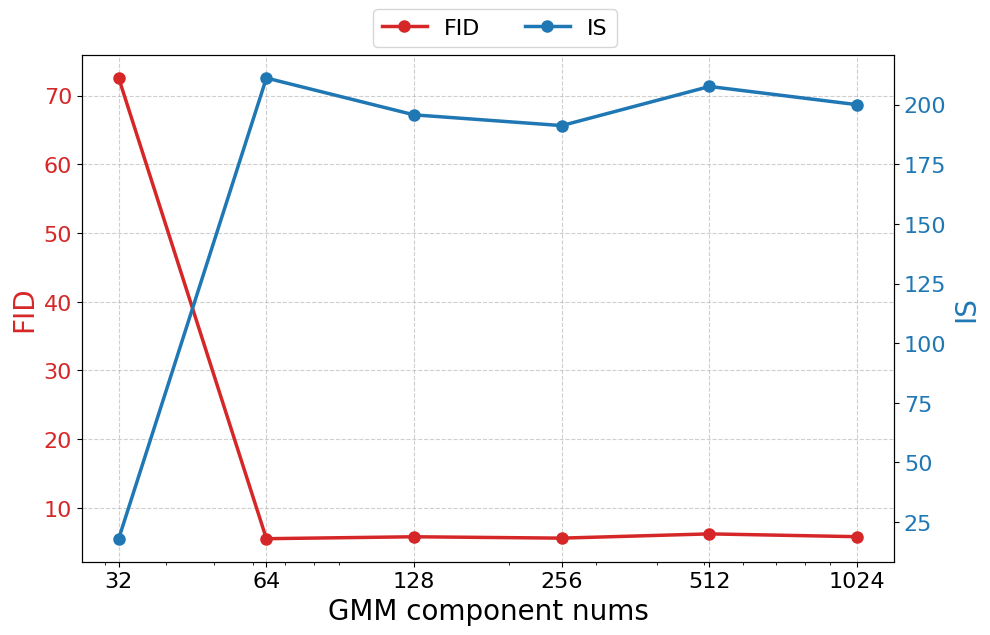}
        \caption{Impact of GMM mixture component number}
        \label{fig:plot_a}
    \end{subfigure}
    \hspace{0.03\linewidth}
    \begin{subfigure}[]{0.46\linewidth}
        \centering
        \includegraphics[width=\linewidth]{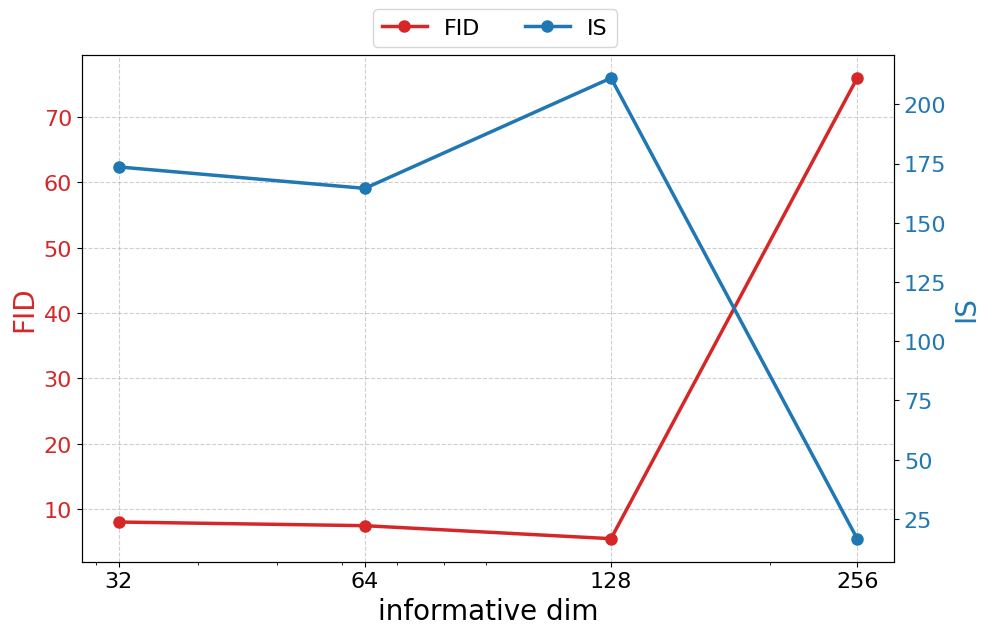} 
        \caption{Impact of informative dimension}
        \label{fig:plot_b}
    \end{subfigure}
    \vspace{-0.5em}
    \caption{\textbf{The ablation study of different properties}.}
    \label{fig:plot}
\end{figure}

\textbf{Impact of GMM Mixture Component Number}.
We analyze the impact of the number of GMM mixtures predicted by the AR models, which reflects the complexity of the approximated distribution. A larger number of mixtures enables the model to represent more complex distributions; however, it also increases sampling difficulty and computational cost during training. As shown in \cref{fig:plot_a}, the FID varies only slightly across different mixture numbers and attains its optimal value at 64 mixtures. Notably, reducing the number further—to 32 mixtures—prevents the model from performing effective dimension reduction, resulting in a significant decline in generation quality. Therefore, we set the number of mixtures to 64 to balance generation quality and training cost.

\textbf{Impact of the Informative Dimension}.
We analyze the impact of the informative dimension, which reflects how information is separated and allocated by the NF models. As shown in \cref{fig:plot_b}, the FID initially decreases as the informative dimension increases and achieves the optimal value at 128. Further increasing the dimension leads to a rise in FID. This phenomenon demonstrates a trade-off: increasing the informative dimension allows capturing more information, but also makes AR modeling and sampling more challenging. Therefore, we set the informative dimension to 128.

\begin{figure}
    \centering
    \includegraphics[width=\linewidth]{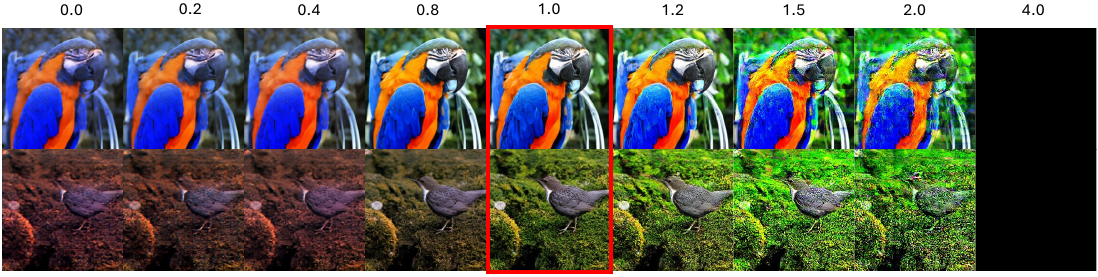}
    \vspace{-2.0em}
    \caption{\textbf{The impact of redundant channels}. The numbers above indicate scaling factors applied to the variance of the shared GMM distribution for redundant channels. Adjusting this variance controls sampling diversity: larger variance yields more diverse, potentially out-of-distribution samples, while smaller variance limits diversity. Visualization results demonstrate that the self-supervised dimension reduction effectively separates structural information from color details.}
    \label{fig:dim_ablation}
\end{figure}

\textbf{Information Separation of Different Dimension Groups}. \label{sec:dim_seperate}
Here, we visualize the information contained in the informative and redundant channels. Specifically, during inference, we first predict all tokens of the informative channels in a token-by-token manner. Subsequently, based on these tokens, we predict a shared GMM distribution for the redundant channels. By adjusting the variance of each Gaussian component in the GMM, different distributions are obtained, from which we sample all tokens of the redundant channels. As shown in \cref{fig:dim_ablation}, reducing the variance causes sampled tokens of redundant channels to concentrate around the means of the Gaussians, resulting in reduced diversity and smoother color regions, while the global structure of the images remains largely unaffected. Conversely, increasing the variance enhances diversity but raises the risk of sampling out-of-distribution values, which can lead to color artifacts or, in extreme cases, failure to generate coherent images. These observations demonstrate that our self-supervised dimension reduction method successfully decouples structural contour information from fine color details.

\begin{table}[!th]
\caption{\textbf{Inference Speed Accelerate}.}
\vspace{-0.5em}
    \centering
    \begin{tabular}{l|cccccc}
    \toprule
      Method & Epochs & FID & IS & \makecell{AR infer. time \\(\% in total)} & \makecell{NF reverse time \\(\% in total)} & Total time \\
       \midrule
               FARMER & 280      & 5.55& 194.63&    0.0500s (22.8\%)    &  0.1689s (77.2\%)  &  0.2189s \\
 w/. One-step Distll. & 280+60   & 5.63& 193.49&    0.0500s (88.2\%)    &  0.0076s (13.4\%) &  0.0567s \\
    \bottomrule
    \end{tabular}
    \label{tab:inference_accelerate}
\end{table}

\textbf{Inference Speed Acceleration}.
As shown in Table~\ref{tab:inference_accelerate}, the baseline FARMER requires 0.2189 seconds per image for inference, where the AR Transformer accounts for 0.0500 seconds and the NF reverse process dominates with 0.1689 seconds. By applying the proposed one-step distillation strategy, the NF reverse time is dramatically reduced from 0.1689 to 0.0076 seconds, yielding a $22\times$ acceleration for this component. Consequently, the total inference time decreases from 0.2189 to 0.0567 seconds per image, nearly a $4\times$ overall speed-up, while maintaining comparable image quality (FID 5.63 vs. 5.55, IS 193.49 vs. 194.63). These results demonstrate that one-step distillation effectively eliminates the sequential bottleneck of the reverse process, enabling FARMER to achieve both high fidelity and efficient generation.

\begin{figure}[H]
    \centering
    \includegraphics[width=\linewidth]{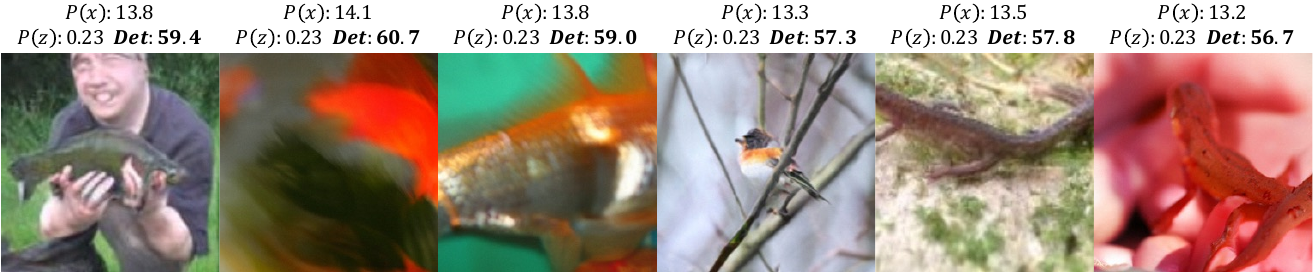}
        \vspace{-2.0em}
    \caption{\textbf{The sample images with abnormal log-determinant values}. High logdet values cause strong compression in parts of the data space, leading to blurred textures and missing fine-scale details in the generated images.}

    \label{fig:det_example}
\end{figure}

\textbf{Impact of Logdet}
As defined in the training objective (see \cref{eq:total_loss3}), the log-determinant (logdet) loss term quantifies the volume change induced by the transformation from the original space to the target latent space. As illustrated in \cref{fig:det_example}, samples with abnormal logdet values often exhibit a blurred appearance and lack fine-grained details. Excessively large logdet values indicate that certain regions of the latent space are strongly compressed in the data space, which can lead to significant errors when reversing the transformation and reconstructing the data. This suggests that maintaining stable logdet values is crucial for high-fidelity and detail-preserving generation.

\section{Conclusion}
We introduce FARMER, a novel generative framework that integrate invertible AF with AR model, enabling end-to-end training directly on raw image pixels. FARMER learns by mapping the data distribution to an distribution  modeled by the AR model and maximizing the negative log-likelihood of the raw images. This design permits both high-quality image synthesis and explicit likelihood estimation. Furthermore, we propose key techniques: a self-supervised dimension reduction to alleviate the complexity of AR modeling/sampling, a resampling-based CFG strategy to enhance image quality, and a one-step distillation scheme to accelerate the inference speed. Through the contributions, FARMER demonstrates competitive performance in image generation relative to pixel-based and latent generative models.
However, beyond the curse of high-dimensionality that we have addressed, two challenges persist in NF–AR,
\ie, (i) dequantization relying on noise injection and (ii) the complications arising from the log-determinant loss. We leave these for future works.

\clearpage

\section{Related Work}

\subsection{Continuous AR}
A common paradigm in autoregressive image generation is to quantize images into discrete tokens~\cite{vqvae2,van2017neural,ramesh2021zero,chang2022maskgit,llamagen,yu2022scaling,lee2022autoregressive} and train autoregressive models over them, as exemplified by LlamaGen~\cite{llamagen}, Janus-Pro~\cite{janus_pro}, and SimpleAR~\cite{wang2025simplear}. However, this design suffers from a key bottleneck: quantization inevitably introduces information loss, which limits the fidelity of generated images~\cite{tschannen2023givt,mar,han2024infinity}.

To address this issue, GIVT~\cite{tschannen2023givt} uses continuous latents obtained from a VAE to encode images and trains an AR model to predict GMM parameters for approximating token distributions. ARINAR~\cite{arinar} further predicts GMM parameters of each token in Gaussian-to-Gaussian paradigm. Since GMMs have limited expressive power, Tschannen et al. further introduce a NF to transform GMM samples into tokens, thereby improving generation quality. Jetformer~\cite{tschannen2024jetformer} goes one step further by discarding the VAE and directly training AR and NF models in the pixel space.

Another line of work explores continuous token modeling by combining AR with diffusion models. In MAR~\cite{mar}, the AR backbone first outputs a conditioning vector for each token, and the diffusion head then generates the next tokens conditioned on this vector. Building on this idea, several other continuous-token approaches have been proposed~\cite{ren2024flowar,harmon,cam,himar,xar,ke2025hyperspherical,deng2024autoregressive,yu2025frequency,liao2025mogao,wu2025omnigen2,ma2025janusflow,bagel,hang2025fast,team2025nextstep}. For example, FlowAR~\cite{ren2024flowar} employs a VAR~\cite{var} backbone with flow matching as the generative head; Hi-MAR~\cite{himar} pivots on low-resolution image tokens
to trigger hierarchical autoregressive modeling in a multi-phase manner; xAR~\cite{xar} autoregressively generates next groups of tokens through flow matching. Although diffusion-based methods are effective at sampling continuous tokens, they require iterative noise-to-token denoising, which limits the model ability to perceive and understand images. In contrast, our model directly fits the token distribution without relying on noise sampling.

\subsection{Autoregressive Normalizing Flow}
Normalizing flows (NF)~\cite{vnf,papamakarios2021normalizing,kobyzev2020normalizing,nice,nvp,glow,van2016pixel,draxler2024free,mathieu2020riemannian,giaquinto2020gradient,draxler2024universality,mate2022flowification} provide a powerful framework for density estimation, visual generation, and text generation~\cite{tarflowllm}, via invertible transformations, enabling exact likelihood computation and efficient sampling. However, the representational capacity of NFs is limited by the expressiveness of these invertible transformations. To address this limitation, autoregressive normalizing flows have been proposed, where each token is transformed conditioned on previous tokens. There has been a long line of work on autoregressive normalizing flows, with representative approaches including IAF~\cite{iaf}, MAF~\cite{maf}, neural autoregressive flows~\cite{huang2018neural}, and T-NAF~\cite{t_naf}. More recently, the resurgence of NFs has attracted renewed interest. TARFlow~\cite{zhai2024normalizing} leverages causal Transformers and simplifies the log-determinant term in the loss function, leading to notable improvements in generation quality. STARFlow extends TARFlow into the VAE latent space and demonstrates that continuous AR flows can deliver competitive generative performance. Meanwhile, JetFormer~\cite{tschannen2024jetformer} integrates Jet~\cite{kolesnikov2024jet} to enable fully end-to-end continuous AR modeling directly over raw image pixels.

\bibliographystyle{plainnat}
\bibliography{main}

\clearpage

\beginappendix

\section{Discussions}
\subsection{FARMER reduces to Autoregressive Flow when $K=1$}
\label{app:k=1}
When the number of components ($K$) in the Gaussian Mixture Model (GMM) predicted by FARMER is set to one ($K=1$), FARMER reduces to an Autoregressive Flow (AF). In this case, each token $z_i$ in the sequence is modeled by a conditional Gaussian distribution, where the mean and variance are functions of the preceding tokens $z_{<i}$.
The optimization objective for each token becomes:
\begin{equation}
    \log p(z_i | z_{<i}) = \log \left( \mathcal{N}(z_i; \mu(z_{<i}), \sigma^2(z_{<i})) \right)
\end{equation}
This can be further expressed as:
\begin{equation}
    \log p(z_i | z_{<i}) = \log \left[ \mathcal{N}\left( \frac{z_i - \mu(z_{<i})}{\sigma(z_{<i})}; 0, I_d \right) \left\lvert 
    \frac{\partial\, [z_i - \mu(z_{<i})]/\sigma(z_{<i})}{\partial z_i} \right\rvert \right],
\end{equation}
where $\mathcal{N}(\cdot; 0, I_d)$ denotes the standard normal density, and the second term inside the log corresponds to the change of variables formula (the volume correction by the Jacobian determinant).

Expanding the log yields two components:
\begin{equation}
    \log p(z_i | z_{<i}) = \log \mathcal{N} \left( \frac{z_i - \mu(z_{<i})}{\sigma(z_{<i})}; 0, I_d \right) + \log \left| \frac{1}{\sigma(z_{<i})} \right|,
\end{equation}
$z_i$ is transformed to new token $\frac{z_i - \mu(z_{<i})}{\sigma(z_{<i})}$ by the predicted results $(\mu(z_{<i}),\sigma(z_{<i}))$ of the AR model conditioned on preceding tokens $z_{<i}$, and this transformation is invertible; the first term is the log-likelihood of new token $z'_i$ under the standard Gaussian distribution, and the second term is the log-determinant of the Jacobian of the affine transformation. Thus, the AR model can be considered as the last block of AFs.

This confirms that when $K=1$, FARMER reduces to an Autoregressive Flow.

\subsection{Resample-based CFG}
\label{app:cfg}
While the main text (\cref{sec:cfg}) outlines the proposed Resampling-based Classifier-Free Guidance (CFG) method, we further elaborate on additional tunable parameters that enhance generation quality and control diversity. Each of the three stages---\emph{Propose}, \emph{Weigh}, and \emph{Resample}---introduces dedicated temperature coefficients and sampling numbers that can be adjusted.

\textbf{Propose stage.}
In the proposal step, candidate samples are drawn from the conditional GMM $p_c(z_i)$ and the unconditional GMM $p_u(z_i)$.  
To control the diversity at this stage, we introduce two distinct temperature coefficients:
\begin{itemize}
    \item \emph{Weight temperature} $T_{\pi}$: applied multiplicatively to the mixture weights $\pi_k(z_{<i})$ of the GMM components before normalization. This modulates the relative selection probability among Gaussian components.
    \item \emph{Variance temperature} $T_{\sigma}$: applied multiplicatively to the variance $\sigma_k(z_{<i})$ of each Gaussian component, scaling the spread of proposals.
\end{itemize}
Additionally, the number of samples drawn from $p_c$ and $p_u$ can differ; we denote these by $s_c$ and $s_u$. This allows balancing between strongly conditioned proposals and broader unconditional exploration.

\textbf{Weigh stage.}
Given candidate samples $z_{i,j}$, their importance weights are computed as:
\[
    \log \omega_{i,j} = w \cdot \big( \log p_c(z_{i,j}; T_{\pi,v}, T_{\sigma,v}) - \log p_u(z_{i,j}; T_{\pi,v}, T_{\sigma,v}) \big),
\]
where $T_{\pi,v}$ and $T_{\sigma,v}$ are temperature coefficients for the evaluation distributions in this stage (not necessarily equal to those used in the \emph{Propose} stage).  
These temperatures control the sharpness or smoothness of the scoring in the log-probability space.

\textbf{Resample stage.}
Finally, the normalized weights $\omega_{i,j}$ define a categorical distribution. To further modulate selection sharpness, we introduce a \emph{resampling temperature} $T_s$ applied uniformly to all log-weights before normalization:
\[
    p_{\text{final}}(z_{i,j}) \propto \exp\big( \log \omega_{i,j} * T_s \big).
\]
Higher $T_s$ emphasizes high-weight proposals, while lower $T_s$ encourages diversity.

\textbf{Summary table of parameter choices.}
Table~\ref{tab:cfg_params} summarizes the temperature and sampling configurations used for different model variants evaluated in this work.

\begin{table}[h]
\centering
\caption{Temperature and sampling parameters for Resampling-based CFG in different models.}
\label{tab:cfg_params}
\vspace{-0.5em}
\begin{tabular}{lcccccccc}
\toprule
Model & $T_{\pi}$ & $T_{\sigma}$ & $s_c$ & $s_u$ & $T_{\pi,v}$ & $T_{\sigma,v}$ & $T_s$ & CFG\\
\midrule
FARMER 1.1B (patch 16)  & 1.0  & 0.9 & 5  & 5  & 0.2 & 0.9 & 1.1 & 2.5 \\
FARMER 1.1B (patch 8)   & 1.0  & 1.0 & 5  & 5  & 0.2 & 0.9 & 1.1 & 2.0 \\
FARMER 1.9B (patch 16)  & 1.0  & 0.9 & 5  & 5  & 0.2 & 0.9 & 1.1 & 3.5\\
FARMER 1.9B (patch 8)   & 1.0  & 1.0 & 5  & 5  & 0.1 & 1.0 & 1.1 & 1.5\\
\bottomrule
\end{tabular}
\vspace{-0.5em}
\end{table}

\end{document}